\documentclass[10pt,twocolumn,letterpaper]{article}

\usepackage{cvpr}
\usepackage{times}
\usepackage{epsfig}
\usepackage{graphicx}
\usepackage{amsmath}
\usepackage{amssymb}
\usepackage{algorithm2e}
\usepackage{booktabs, multirow} 
\usepackage{soul}
\usepackage[table]{xcolor} 
\usepackage{changepage,threeparttable} 
\usepackage{caption}

\usepackage[pagebackref=true,breaklinks=true,letterpaper=true,colorlinks,bookmarks=false]{hyperref}

\cvprfinalcopy 

\def\cvprPaperID{9510} 

\ifcvprfinal\fi

\makeatletter
\newcommand{\printfnsymbol}[1]{%
  \textsuperscript{\@fnsymbol{#1}}%
}
\makeatother

\begin{document}

\title{Box Supervised Video Segmentation Proposal Network}

\author{Tanveer Hannan\thanks{Contributed equally and share the first-authorship of this paper.}\\
Ludwig-Maximilians-University\\
Munich, Germany\\
{\tt\small hannan@dbs.ifi.lmu.de}

\and
Rajat Koner\printfnsymbol{1}\\
Ludwig-Maximilians-University\\
Munich, Germany\\
{\tt\small koner@dbs.ifi.lmu.de}

\and
Jonathan Kobold\\
Hensoldt Analytics\\
Munich, Germany\\
{\tt\small Jonathan.Kobold@hensoldt.net}

\and
Matthias Schubert\\
Ludwig-Maximilians-University\\
Munich, Germany\\
{\tt\small schubert@dbs.ifi.lmu.de}
}

\maketitle

\begin{abstract}
Video Object Segmentation (VOS) has been targeted by various fully-supervised and self-supervised approaches. While fully-supervised methods demonstrate excellent results, self-supervised ones, which do not use pixel-level ground truth, attract much attention. However, self-supervised approaches pose a significant performance gap. Box-level annotations provide a balanced compromise between labeling effort and result quality for image segmentation but have not been exploited for the video domain. In this work, we propose a box-supervised video object segmentation proposal network, which takes advantage of intrinsic video properties. Our method incorporates object motion in the following way: first,  motion is computed using a bidirectional temporal difference and a novel bounding box-guided motion compensation. Second, we introduce a novel motion-aware affinity loss that encourages the network to predict positive pixel pairs if they share similar motion and color. The proposed method outperforms the state-of-the-art self-supervised benchmark by 16.4\% and 6.9\% $\mathcal{J}$\&$\mathcal{F}$ score and the majority of fully supervised methods on the DAVIS and Youtube-VOS dataset without imposing network architectural specifications. We provide extensive tests and ablations on the datasets, demonstrating the robustness of our method. Code is available at \url{https://github.com/Tanveer81/BoxVOS.git}
\end{abstract}

\section{Introduction}
Video Object Segmentation (VOS) primarily consists of two stages. First, it separates and segments objects from their surroundings, later propagates them throughout the video sequence. It is a challenging problem as the objects in a video change their appearances due to exposure, rotation, and occlusion. VOS has been solved in two ways. First fully supervised VOS frameworks \cite{cfbi}, that utilize rich frame-wise manual annotation. Second, self-supervised VOS \cite{ CorrFlow}, in contrast to the fully-supervised approach, do not need annotations. Instead, they explore intrinsic video properties like motion, optical flow, or other representations. Self-supervised VOS provides an impressive approach that avoids the need for expensive frame-wise object masks. However, they suffer from a significant performance gap compared to their supervised counterparts.

\begin{figure}[tbh]
\begin{center}
  \includegraphics[width=1.0\linewidth]{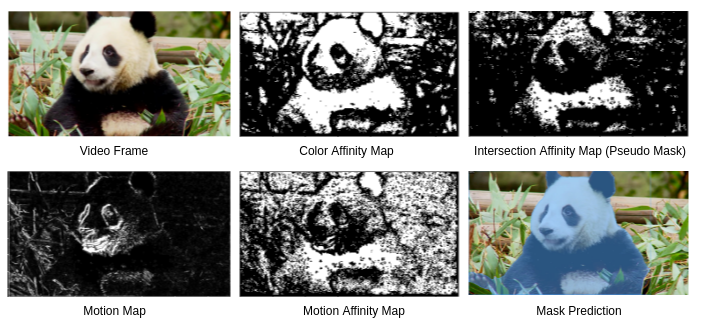}
\end{center}
    \caption{The pair-wise affinity maps are generated based on the frame's color and motion separately. Then, the intersection affinity map(pseudo mask) is created where pixel pairs share similar motion and color characteristics. The intersection improves performance by reducing background noise and generating a sparse, high-precision pseudo mask.}
\label{fig:intro}
\end{figure}

In this work, we have tried to bridge the gap only using box annotation and intrinsic video properties. To the best of our knowledge, we are the first to propose a box-supervised video object segmentation proposal network without using any ground truth mask annotation. The central idea of this work is to distinguish similar regions based on motion and color similarity utilizing our novel motion aware affinity loss. Hence, we propose two new components for the affinity loss: first, train time motion generation using a bounding box guided motion compensation method. Second, we introduce pseudo masks, i.e., an approximation of the ground truth masks as in figure \ref{fig:intro}, derived from the combined color and motion. Our method can be easily integrated into most existing VOS frameworks to generate high-quality mask proposals without modifying the network itself. We achieve competitive performance compared to supervised baselines and significantly outperform other self-supervised approaches.  
We achieve competitive performance compared to supervised baselines and significantly outperform self-supervised approaches on the established DAVIS and Youtube-VOS benchmarks.  

In weakly supervised image segmentation tasks, bounding box supervision \cite{ Lan2021} improves the quality of pseudo mask generation compared to class label supervision \cite{Xploiting2021} with a slight increase in annotation effort. It inspired us to use weak bounding box-based supervision for video data instead of expensive pixel-wise mask annotations. However, foreground objects in a video constantly change their appearance. Thus, separating them using only color information can result in sub-optimal performance. In contrast to color, object motion in a video is an independent yet complementary feature that can help to distinguish the object from the background. We argue that incorporating extended temporal motion information can be exploited for high-quality mask generation. Motion computation is inherently noisy due to global camera and other minute movements. To mitigate this, an affine transformation is employed on the current frame, which aligns the backgrounds of subsequent frames. Moreover, the pixels outside the bounding box are exploited for better background characteristics and transformation matrix computation. The improved alignment ensures that the foreground and moving objects are the primary sources of motion response. 

The generated motion is then fused with color to create the pseudo mask for supervision. The proposed motion aware affinity loss uses box supervision where a pixel pair is located inside a box compared to its surroundings if their motion and color characteristics are similar. It leads to a tight and precise mask computation for a perspective foreground object. For example, in Figure \ref{fig:intro} a panda sitting in a bamboo field, where the panda's facial fur is camouflage or has a similar color compared to the wooden background stem. It poses a considerable challenge to distinguish the object pixel from the background based on color similarity \cite{Lamdouar}. Hence, resulting in erroneous mask computations, which include background pixels. However, the inclusion of motion could alleviate this problem, as it is invariant color. We hypothesize that the intersection of pixels with color and motion similarity will increase foreground pixels' precision and help to compute a tight mask around the object. Figure \ref{fig:intro} shows the generation of affinity maps.

Furthermore, incorporating our novel motion-aware affinity loss requires no architectural changes. Finally, the contribution of our work can be summarized as follows:
\begin{itemize}
    \item We are the first to propose a video segmentation proposal network that employs weak bounding box supervision and does not require any ground-truth mask. Our method is compatible with most of the existing VOS frameworks and does not require any changes to their parameters.    
    \item An improved motion compensation technique is proposed that effectively utilizes the available bounding box coordinates for reducing global camera movement.
    \item A novel motion aware affinity loss is introduced for frame-level segmentation with a pseudo mask generated from pair-wise color and motion similarity.  
    \item We have conducted extensive experiments on multiple datasets and ablation studies. Our method achieved absolute 16.4\% and 6.9\% J\&F improvements on the DAVIS and Youtube-VOS datasets.
\end{itemize}
\section{Related Work} 
Supervised learning immensely accelerated various computer vision tasks like object detection and segmentation \cite{carion2020end}, relation prediction \cite{koner2020relation,koner2021scenes} and several downstream task like VQA \cite{hildebrandt2020scene,koner2021graphhopper} and others \cite{koner2021oodformer}. However, annotating segmentation masks for an object is much more expensive than a box or object-label annotation for object detection or relation prediction. It led to the exploration of image instance segmentation without using an expensive segmentation mask.
\begin{figure*}[!htp]
\centering
\includegraphics[width=0.95\linewidth]{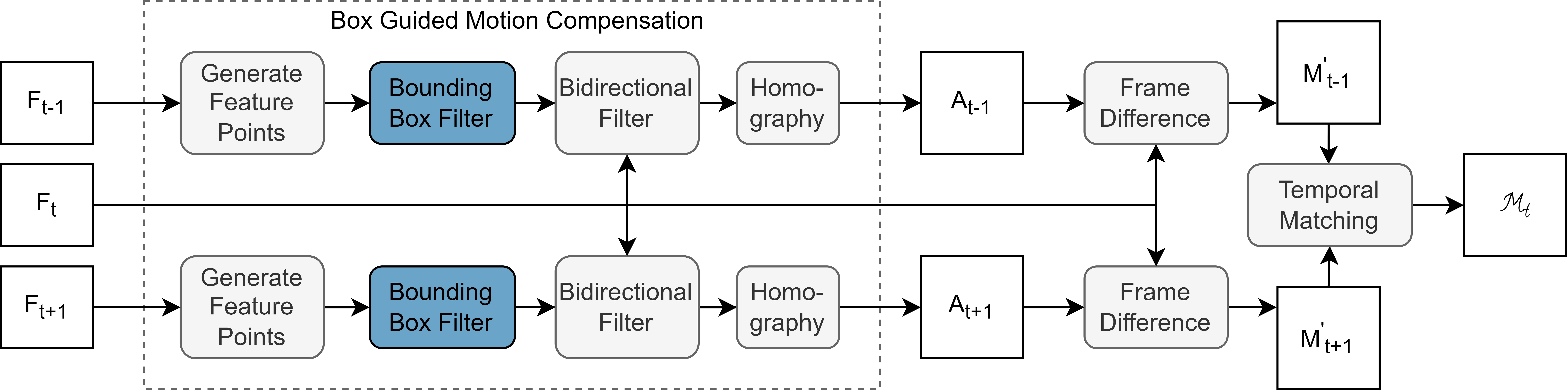}
\caption{\textbf{Proposed motion calculation pipeline.} The first step, is box-guided motion compensation to reduce camera movement. At second, the motion maps from the previous to the current frame ( $M'_{t-1}$) and from the current to the next frame ($M'_{t+1}$) are derived by the frame difference method. Finally, we combine $M^{'}_{t-1}$ and $M^{'}_{t+1}$ (Temporal matching) to generate superior motion map ($\mathcal{M}_t$) for current frame ($F_t$).}
\label{fig:pipeline}
\end{figure*}
\paragraph{Box Supervised Image Instance Segmentation} 
The primary challenge for exploring instance segmentation from images without using ground truth masks is generation pseudo masks. There are four approaches like key-points \cite{keypoint1,keyscribble}, scribbles \cite{keyscribble,scribble1}, class labels \cite{AhnDGIST,imgtag1,imgtag4}, and bounding boxes \cite{Xploiting2021,Lee,Song2019,Kulharia2020,Lempitsky2009,Khoreva2017} address this.  Existing methods accomplish this using either classical cut-based methods \cite{Kulharia2020, Khoreva2017} or by leveraging neural network activation maps \cite{class_sup}. Among these, the state-of-the-art box supervised instance segmentation methods are BoxInst \cite{Tian2020} and DiscoBox \cite{Lan2021}. BoxInst uses the color of nearby pixels to cluster them and forces the neural network to predict a similar class for the pixels. The pairwise loss and projection loss aid the network in learning segmentation in a weakly supervised manner. DiscoBox, on the other hand, generates pseudo labels through multi-instance learning \cite{Hsu2019}. However, DiscoBox is not directly applicable to VOS as it requires class-level labels to train the correspondence matching. This paper explores box supervision on video data which allows our model to exploit inter-frame dependency for VOS.
\paragraph{Motion Generation}
Usually, the motion map in a video is calculated by subtracting the background from two consecutive frames ~\cite{Ellenfeld}. In the case of global camera movement, motion compensation \cite{hartley_zisserman_2004} is used. To reduce noise, forward-backward sparse optical flow \cite{Wan2014} or three-frame difference \cite{Sommer2016,Chen2012} is used. Our method employs multiple filters and incorporates bounding box coordinates to improve the motion calculation further.

\paragraph{Segmentation using motion}
Existing works utilize optical flow to approximate the motion of objects. For example, FlowIRN \cite{Liu} uses class activation maps \cite{irn} and dense optical flow \cite{ddflow} to generate pseudo supervision. In Motion Grouping \cite{YangGrouping}, dense optical flow \cite{raft,pwcnet} is used to cluster pixels into foreground and background. In Deep Fusion\cite{Ellenfeld}, a motion map calculated from sparse optical flow \cite{kanade} is used as network input. Dense optical flow \cite{pwcnet} is used to detect camouflaged animals in \cite{Lamdouar}. The optical flow generated motion map is used as input in the mentioned works. In contrast, our work investigates the possibility of training the segmentation network using motion only during training time and generating pseudo labels by fusing it with the color of objects.

\paragraph{Semi-Supervised VOS} is a recent challenge introduced by DAVIS \cite{davis17} and adopted by Youtube-VOS \cite{yvos2018}. The term ``semi-supervised" does not refer to the level of supervision used during training. Instead, it only refers to the availability of annotation masks for the first frames of test videos. The challenge is to generate segmentation in the subsequent frames from the given first frame. Some methods \cite{Xie, stm, cfbi, premvos} solve the task in a fully-supervised setup, where segmentation masks are used for supervision.
On the contrary \cite{vid_col, CorrFlow, lu2020learning}, trained their models in a self-supervised way like by exploiting deep optical flow to generate segmentation. Although self-supervised methods explore a promising direction, they suffer a significant performance gap with fully supervised ones. Our approach tries to bridge this gap using only box annotation. 
\section{Method}
\label{sec:method}
We first introduce our novel motion compensation and generation pipeline, section \ref{sec:motion_calc}, and the motion aware affinity loss, in section \ref{sec:loss}.
The segmentation network and object tracker used in this paper are explained in section \ref{sec:network}.

\subsection{Proposed Motion Calculation Pipeline}
\label{sec:motion_calc}
Unlike in a single image, we can assess object movement in a video. In addition to the object's motion, a moving camera can add apparent movement to a video because of the shifting point of view. In the presence of this global camera motion, the computation of motion maps is an intricate task and subject to ongoing research  \cite{ Yu2019}. Figure \ref{fig:pipeline}, depicts our novel motion computation pipeline that mitigates the earlier effect through refinement of motion using multiple filters and given bounding boxes. 
\begin{figure}[h]
    \centering
    \includegraphics[width=1\linewidth]{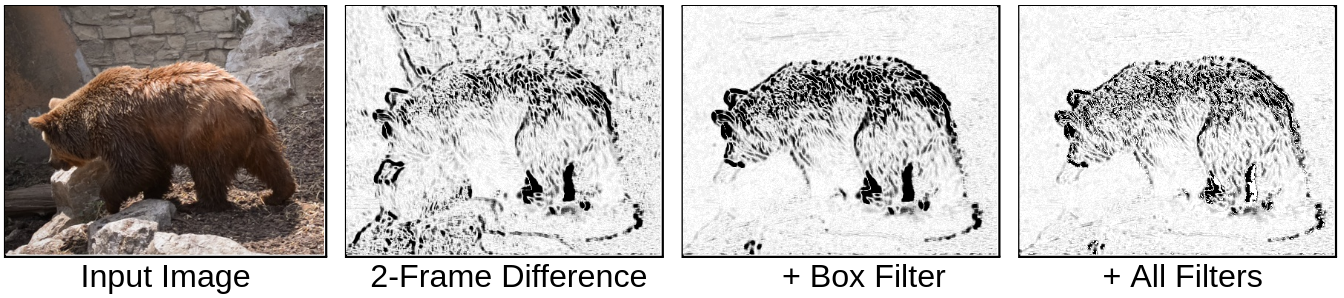}
    \caption{Improvement of motion map with additional filters.}
    \label{fig:compensation_example}
\end{figure}

\paragraph{Box Guided Motion Compensation:} 
In the case of a stationary camera, motion can be calculated by background subtraction from consecutive frames. However, in a video captured with a moving camera, the background moves as well. In order to reduce background motion, the previous and current frames must be co-registered, i.e., the backgrounds of those frames must be aligned. 
\begin{figure}[htp]
    \centering
    \includegraphics[width=0.95\linewidth]{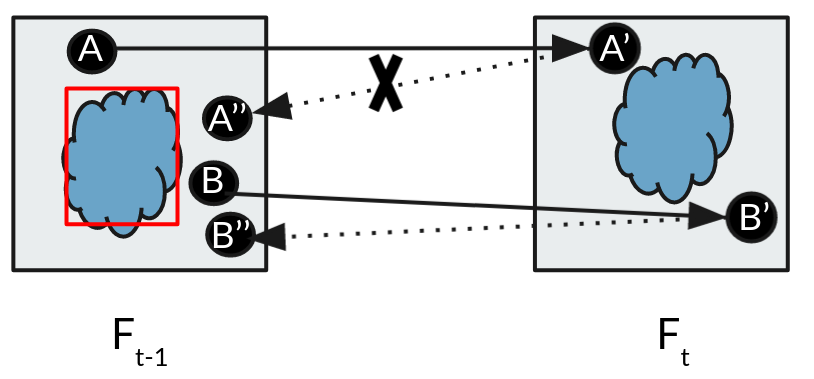}
    \caption{A set of feature points $\{A,B\} \subseteq \mathcal{F}$ are extracted from the previous frame $F_{t-1}$ that lies in region outside the box. 
    Let $\{A^{'}$,$B^{'}\}$ be the forward projection of $\{A,B\}$ and $\{A^{''}$,$B^{''}\}$ be the backward projection of $\{A^{'}$,$B^{'}\}$ obtained from sparse optical flow. Finally, we take the point that represents better background properties or if the original point has less distance from its reverse projected point, in the example its $(B, B^{''})$.}
    \label{fig:compensation}
\end{figure}
At time $t$ the previous frame $F_{t-1}$ is aligned to the current frame $F_t$ with a homography matrix $\mathcal{H}$. The homography matrix $\mathcal{H}$, is the affine transformation between the point pairs from $F_{t-1}$ and $F_t$. $\mathcal{H}$ can be calculated by sampling feature points from $F_{t-1}$ and $F_t$ which are matched with the RANSAC-algorithm.
We use \cite{323794} for extracting feature points. Notably, this co-registration or background alignment works well if these features come from the background, not the foreground. Hence, bounding boxes can be exploited in a weakly supervised setup to sample feature points only from outside the box, we termed it as Box Filter (BF). However, annotations usually only comprise primary objects, background clutters remain a big source of motion. A Bi-Directional Filter (BDF) is employed further to refine the feature points and better background modeling. Figure \ref{fig:compensation}, depicts our BDF, where A and B are two points from frame $F_{t-1}$  along with their forward $\{A^{'}$, $B^{'}\}$ and backward $\{A^{''}$, $B^{''}\}$  projection obtained from sparse optical flow\cite{kanade}. Finally, we remove points that have an L2 distance $D_A$ between the original point (i.e., A) and backward projected point (i.e.,$A^{''}$) greater than $\tau$ where,

\begin{equation}
\label{eq:boxmotion}
    \tau = median(\{D_P|P \in \text{feature points}\}).
\end{equation}

 Hence, the homography matrix $\mathcal{H}$, obtained from these remaining points is used to transform $F_{t-1}$ to the co-registered frame $A_{t-1}$ as shown in Figure \ref{fig:pipeline}. Finally, the motion from two consecutive frames is obtained as $M'_{t-1} = |F_t-A_{t-1}|$ corresponds to a background subtraction. Figure \ref{fig:compensation_example} shows the improvement of the motion map for moving objects with the proposed feature point filtering.  

\paragraph{Temporal Matching:} Often, indiscriminate movement between two consecutive frames causes low-quality motion maps due to a short temporal span or erroneous optical flow. We incorporate motion from both forward and backward temporal directions to alleviate these problems and merge them.    
Here, the next frame's motion map ($M^{'}_{t+1}$) from $F_t$ to $F_{t+1}$ is computed similarly to the earlier two consecutive frames difference. Finally, the motion map $\mathcal{M}_t$ with temporal bidirectional matching is obtained as

\begin{equation}
\mathcal{M}_t=\left\{\begin{array}{cc}
M^{'}_{t-1} & \text { if } M^{'}_{t-1} - M^{'}_{t+1} <\tau_2 \\
0, & \text { else }
\end{array}\right.
\label{eqn:three_frame_diff}
\end{equation}
where $\tau_2$ is a predefined threshold. Here the motion response of a certain pixel in $M^{'}_{t-1}$ is filtered out to zero if it is greater than that of $M^{'}_{t+1}$ up to the threshold $\tau_2$. 

\subsection{Motion Aware Affinity loss}
\label{sec:loss}
The generated motion is combined with color to create a pseudo mask approximating the ground truth mask. However, the generated pseudo mask is inherently noisy because of impurities in motion and color. Thus, penalizing the network with global pixel-wise cross-entropy loss would be erroneous. Instead, we can classify pairs of pixels in close vicinity by looking at their color and motion. The idea is to define pixel pairs as positive, i.e., they belong to the same class if they share similar colors and motion in a local neighborhood. Finally, an affinity loss is employed to assign the same class for each positive pixel-pair if they share a similar pair-wise similarity. The mispredicted positive pairs from the segmentation network contribute to the overall loss.

\begin{figure}[!h]
    \centering
    \includegraphics[width=0.95\linewidth]{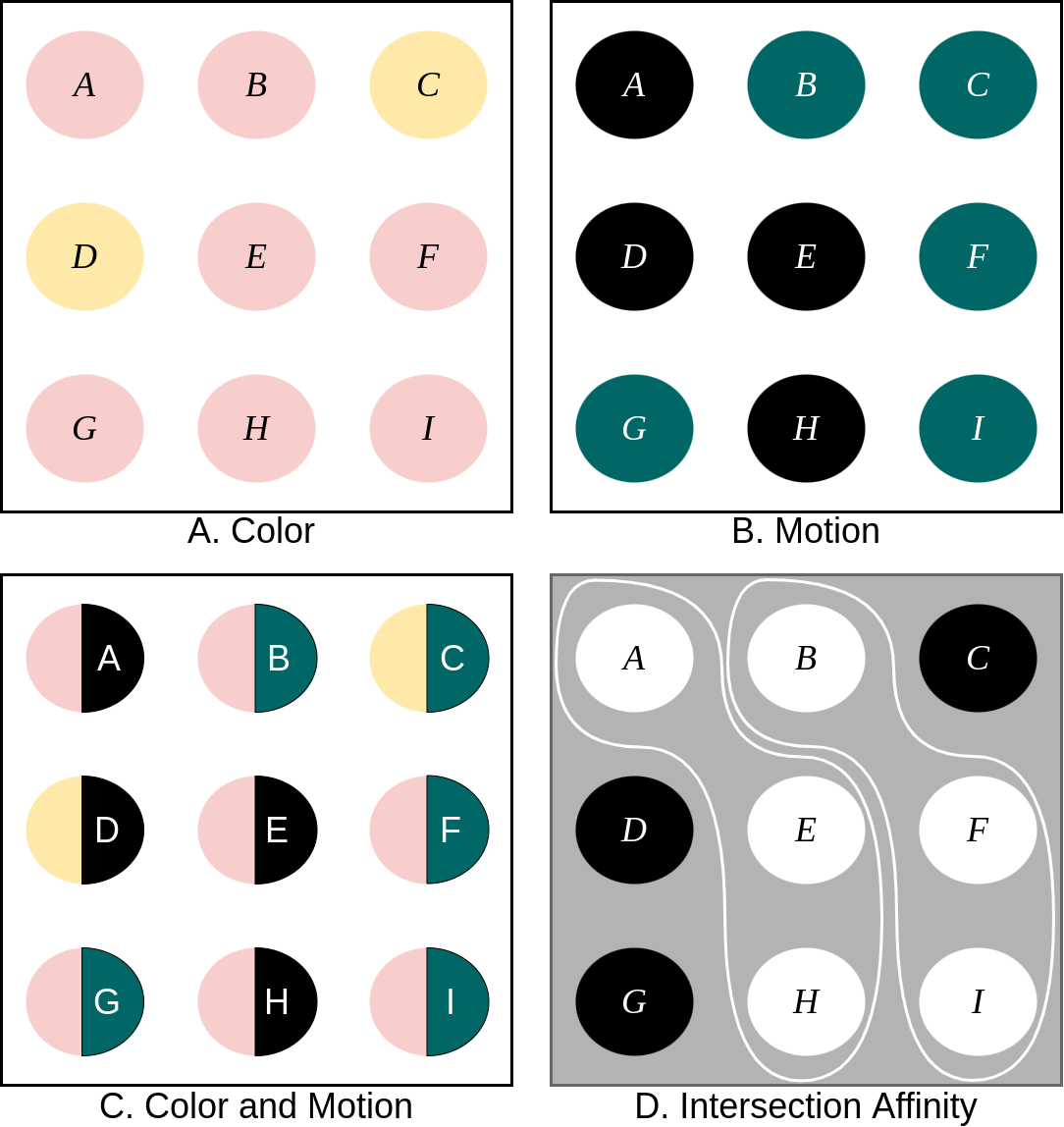}
    \caption{An illustrative example of the motion aware affinity loss in a 3x3 pixel grid of an image. In Fig. A and B pixels with a similar color/motion are indicated by the same color. Fig. C combines the color(left) and motion(right) properties from Fig. A and B. Fig. D shows the final affinity map. Groups of pixels with the same color/motion properties are marked by the white boundary ( e.g., $\{A,E,H\}$, $\{B,F,I\}$). With pairwise pixel matching, each of these groups gets clustered into the same class during training.}
    \label{fig:intersect_aff_map}
\end{figure}

Figure \ref{fig:intersect_aff_map} shows the computation of our proposed loss in a local neighborhood of $3\times3$. Consider pixels $A$ and $B$ and their corresponding motion values are $\mathcal{M_{A}}$ and $\mathcal{M_{B}}$. $\mathcal{C_{A}}$ and $\mathcal{C_{B}}$ are their color values. The motion and color similarities (the higher the value the more similar it is)  between these pixels are $\mathcal{\psi_{A,B}} \in \, ]0,1]$, and $\mathcal{\phi_{A,B}}\in \, ]0,1]$. These similarities can be expressed as 

\begin{equation}
    \begin{aligned}
        \mathcal{\psi_{A,B}} = exp(-\|\mathcal{M_{A} - M_{B}\| * \eta})\\
        \mathcal{\phi_{A,B}} = exp(-\|\mathcal{C_{A} - C_{B}\| * \eta})
    \end{aligned}
    \label{eq:sim}
\end{equation}

where $\eta$ is a hyper-parameter. We hypothesize that two pixels with similar color and motion are likely to be in the same class. For example, the pixel pair $(A, E)$ in Figure \ref{fig:intersect_aff_map}, has similar color and motion, indicating that both pixels belong to the same class. The pixel pair $(A, B)$, on the other hand, has a high color similarity but a low motion similarity, so they should be classified into different classes. Finally, for a pair of pixels their motion ($\Psi_\mathcal{A, B}$) and color ($\Phi_\mathcal{A, B}$) affinity can be defined as, 

\begin{equation}
    \begin{aligned}
        \begin{array}{cc}
        \Psi_\mathcal{A,B}=
        \left\{ \begin{array}{ll}
            1 &\mathcal{\psi_{A,B}} > \tau_m\ \\
            0 &\text{else}
        \end{array} \right. ,&
        \Phi_\mathcal{A,B}=
        \left\{ \begin{array}{ll}
            1 &\mathcal{\phi_{A,B}} > \tau_c\ \\
            0 &\text{else}
        \end{array} \right.
        \end{array}
        \end{aligned}
        \label{eq:aff_mc}
    \end{equation}

 where $\tau_m $ and $\tau_c$ are the motion and color similarity threshold. 
 The proposed loss is calculated only for pixel-pairs with high similarity, i.e., positive affinity. It essentially means the network is trained to predict segmentation with the correct pixel pairs. Hence, constructing high precision positive pixel pairs from motion and color information can significantly boost the network performance. Thus, an affinity ($\mathcal{I_{A,B}}$) between two pixel $(A,B)$ is defined by  

\begin{equation}
\mathcal{I_{A,B}}=\left\{\begin{array}{cc}
1 & \text { if } \Psi_\mathcal{A,B} = \Phi_\mathcal{A,B} \\
0, & \text { else }
\end{array}\right.
\label{eq:aff_i}
\end{equation}

Eq. \ref{eq:aff_i}, effectively reduces the noise from each set. A detailed experiment and discussion on the selection of combined motion and color-based affinity can be found in Table \ref{tab:ablation_fusion}.

Let the confidence score of the segmentation network for pixels $A$ and $B$ being foreground be $\rho_{\mathcal{A}}$ and $\rho_{\mathcal{B}}$. Then the confidence of these two pixels being predicted as the same class (either foreground or background) is $\rho_{\mathcal{A,B}}$,

\begin{equation}
    \rho_{\mathcal{A,B} } = \rho_\mathcal{A} * \rho_\mathcal{B} + (1-\rho_\mathcal{A}) * (1-\rho_\mathcal{B}),
    \label{eq:same_class}
\end{equation}
 where the confidence of the pixel pair being foreground is $\rho_\mathcal{A} * \rho_\mathcal{B}$, or being background is  $(1-\rho_\mathcal{A}) * (1-\rho_\mathcal{B})$. Let $\mathcal{P}$ be the set of pixel pairs where at least one pixel falls inside the bounding box. For a pixel pair A and B, our motion aware affinity loss, $\mathcal{L_{A,B}}$ , is calculated as 

\begin{equation}
    \mathcal{L_{A,B}} = -\mathcal{I_{A,B}} * log(\rho_{\mathcal{A,B}}).
    \label{eq:pair_loss}
\end{equation}

Finally, for all pairs of pixels in $\mathcal{P}$ we calculate the loss as, 

\begin{equation}
    \mathcal{L_{\textit{affinity}}} = \sum_{\mathcal{A,B} \in \mathcal{P}}  \mathcal{L_{A,B}}.
    \label{eq:all_loss}
\end{equation}

\subsection{Segmentation Architecture} 
\label{sec:network}
\paragraph{Proposal Generation}
CondInst\cite{tian2020conditional} is our default segmentation network. It is a Region-Of-Interest(RoI) free segmentation network with limited parameters and efficient computation. The core of CondInst is built on top of the Feature Pyramid Network (FPN) \cite{lin2017feature} and FCOS \cite{tian2019fcos}. FPN employs multi-scale feature maps for detecting objects from various scales. 
FCOS is a proposal-free, anchor-free, fully convolutional one-stage object detector. Instead of assigning multiple anchor boxes\cite{ren2015faster} for each feature point produced from FPN, it predicts only one box. In addition, a center-ness weighting scheme was introduced to filter the boxes far away from the center. Unlike traditional networks \cite{maskrcnn} that define instances with individual boxes and ROI cropping or pooling, CondInst proposes an instance-aware mask head with conditional convolution. It dynamically generates instance-specific convolutional filter parameters. This conditional convolution learns instance properties without a complicated object detection pipeline. 
In this work, the fully supervised mask loss of CondInst is replaced with a projection loss\cite{Tian2020} and the proposed motion aware affinity loss. Projection loss is a specialized 1d intersection-over-union (IoU) loss calculated separately for the height and width of the segmentation mask.

\paragraph{Tracking}
\label{tracking}
As part of the VOS evaluation method, all object proposals need to be tracked throughout the sequence. In order to track generated object proposal between consecutive frames we employed both optical flow \cite{ilg2017flownet} and object re-identification \cite{li2017video} network similar to \cite{premvos}. In semi-supervised VOS evaluation \cite{davis17,yvos2018} ground-truth proposals from the first frame are matched with the next frame proposals using a bipartite matching mechanism. The bipartite matching utilizes a score from IoU overlap wrapped with the optical flow and re-identification score.  The generated proposals for consecutive frames are matched to the previous frame similarly.   

\section{Experiments}
We conduct our experiments on the two most popular video object segmentation datasets, DAVIS-2017 \cite{davis17}, and Youtube-VOS \cite{Yang2019}. Unless otherwise specified, we have only used bounding box supervision for training, and $\mathcal{J}$\&$\mathcal{F}$ score \cite{davis17} for evaluation throughout our experiments and ablation studies.
We computed the bounding box from the segmentation mask as explicit bounding boxes are not given in both datasets. During test-time, we generate frame-wise proposals and track them throughout the sequence in semi-supervised VOS evaluation settings. We compare our new method with the color-only, box supervised, state-of-the-art image segmentation network BoxInst \cite{Tian2020}. For a fair comparison, BoxInst also is based on a CondInst backbone 
and is trained with the same data sets and default parameters. The baseline network is trained with only color information. \\
Details of all the hyperparameters can be found in the appendix.

\subsection{Dataset and Metrics}
\textbf{DAVIS-2017} is one of the most widely used VOS datasets where every frame is annotated with a ground truth mask. A predefined training and validation split consist of 60 and 30 videos. \\
\textbf{Youtube-VOS} was recently proposed as one of the largest video object segmentation datasets. It contains 3,471 training and 507 validation videos with 94 different object categories. Training data contains ground truth mask annotations for every fifth frame, while validation contains only for the first frame. 

\textbf{$\mathcal{J}$ \& $\mathcal{F}$ Score } the Jaccard index $\mathcal{J}$ is the intersection of prediction and ground truth mask over their union. It indicates the region accuracy of the prediction. The contour accuracy $\mathcal{F}$ calculates the f1-score of the boundary pixels with a bipartite matching algorithm. The mean of these $\mathcal{J}$ \& $\mathcal{F}$ is reported as in the DAVIS 2017 \cite{davis17}, for both DAVIS and Youtube-VOS.

\textbf{mAP} the mean average precision is the area under the precision-recall curve where the IoU threshold determines the positive and negative examples. We use the implementation from COCO \cite{coco} for the frame-level evaluation in the ablation study.

\subsection{Results} 
We compare our method from two perspectives. The first one is the comparison with our baseline network to demonstrate how motion solely improves the performance. The second one is the comparative evaluation of our method with fully- and self-supervised VOS methods. We use the ResNet-101 \cite{he2016deep} backbone for the comparison with other state-of-the-art methods.

\label{sec:result}
\begin{figure}[h]
    \centering
    \includegraphics[width=\linewidth]{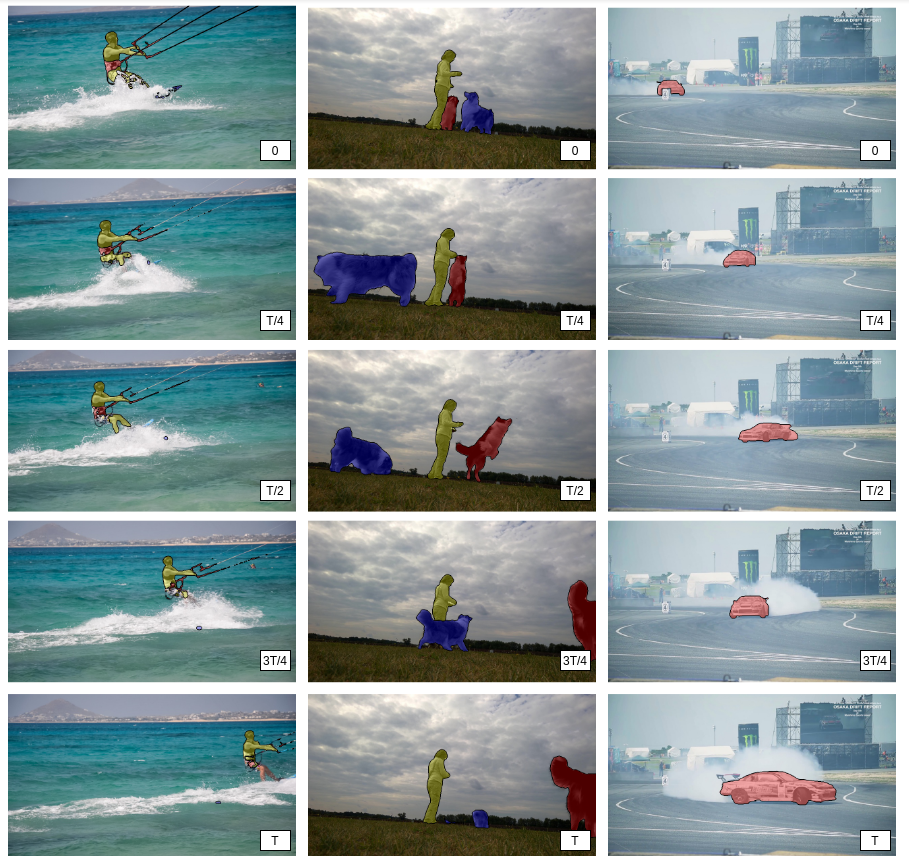}
    \caption{Predictions of our method on the DAVIS data-set. Due to semi-supervised evaluation, the first-frame(0th) represents the ground truth annotation mask.}
    \label{fig:video_sequence}\vspace{-1em}
\end{figure}

\paragraph{Comparison with Box Supervised Color only Baseline} 
In this experiment, we employ our baseline network and train without motion information. The baseline network's architecture and other parameters remain the same as ours. The generated framewise proposals are tracked similarly to those as mentioned in section \ref{tracking}. Both datasets are used to train the baseline for comparability. Table \ref{tab:davis_sota} and Table \ref{tab:uvos_sota}, demonstrate that our method outperforms the color only baseline by 3.6\% on the DAVIS and 5.3\% on the YouTube-VOS dataset. The results demonstrate the advantages of our method and the exploration of motion cues alongside color for VOS.

\paragraph{Comparison with State-of-the-Art VOS methods}
Table \ref{tab:davis_sota}, shows our performance on the DAVIS-2017 validation set.  We significantly outperform the self-supervised benchmark by 16.4\% $\mathcal{J} \& \mathcal{F}$ score on DAVIS. Furthermore, we demonstrate superior performance compared to the most fully supervised approaches and achieved a competitive performance with the top-performing models. Our performance is an indicator that box-guided exploration of simultaneous video cues like motion or color could lead to a potent VOS architecture comparable to fully supervised ones without using ground truth masks. With less expensive box label annotation, we significantly reduce the performance gap with fully-supervised methods that use expensive ground truth mask annotation. Figure \ref{fig:video_sequence} shows some qualitative examples. We also evaluated our approach on the YouTube-VOS dataset \cite{Yang2019} as shown in Table \ref{tab:uvos_sota}. We achieve 53.5 $\mathcal{J} \& \mathcal{F}$ score, which is better than the self-supervised methods and gives competitive results in comparison with fully supervised methods \cite{osmn,msk,rgmp,onavos}.

\begin{table}[htp]
\center
\scriptsize
\begin{tabular}
{cccccc} \toprule
Method & Sup. & $\mathcal{J} \& \mathcal{F}$ & $\mathcal{J}$ & $\mathcal{F}$ \\ \midrule \midrule

Vid. Color.\cite{vid_col}  & Self & $34.0$ & $34.6$ & $32.7$ & \\
CycleTime\cite{cycletime}  & Self & $48.7$ & $46.4$ & $50.0$ & \\
CorrFlow\cite{CorrFlow}  & Self & $50.3$ & $48.4$ & $52.2$ & \\
UVC\cite{uvc} & Self & $59.5$ & $57.7$ & $61.3$ & \\
RPM-Net\cite{rpmnet} & Self & $41.6$ & $41.0$ & $42.2$ & \\
Mug\cite{lu2020learning}  & Self & $56.1$ & $54.0$ & $58.2$ & \\
\midrule

BoxInst\cite{Tian2020} $\dagger$ & \text{Box} & $68.9$ & $68.2$ & $69.6$ & \\
\textbf{Ours} & \textbf{Box} & $\textbf{72.5}$ & $\textbf{71.5}$ & $\textbf{73.5}$ \\ \midrule

OSMN\cite{osmn} & Full & $54.8$ & $52.5$ & $57.1$ & \\
OSVOS\cite{osvos}  & Full & $60.3$ & $56.6$ & $63.9$ & \\
SiamMask\cite{siammask} & Full & $56.4$ & $54.3$ & $58.5$ & \\
OSVOS-S\cite{osvos-s}  & Full & $68.0$ & $64.7$ & $71.3$ & \\
GC\cite{gc} & Full & $71.4$ & $69.3$ & $73.5$ & \\
\textcolor{blue}{FEELVOS\cite{feelvos}} & \textcolor{blue}{Full} & \textcolor{blue}{$71.5$} & \textcolor{blue}{$69.1$} & \textcolor{blue}{$74.0$} & \\
AFB-URR\cite{afburr} & Full & $74.6$ & $73.0$ & $76.1$ & \\
\textcolor{violet}{PReMVOS\cite{premvos}} & \textcolor{violet}{Full} & \textcolor{violet}{$77.8$} & \textcolor{violet}{$73.9$} & \textcolor{violet}{$81.7$} & \\
CFBI\cite{cfbi} & Full & $81.9$ & $79.1$ & $84.6$ & \\
STM\cite{stm} & Full & $81.8$ & $79.2$ & $84.3$ & \\
\textbf{RMNet\cite{Xie}} & \textbf{Full} & $\mathbf{8 3 . 5}$ & $\mathbf{8 1 . 0}$ & $\mathbf{8 6 . 0}$ & \\ \hline

\end{tabular}
\caption{ State-of-the-art comparison on DAVIS \cite{davis17} validation set. $\dagger$ denotes our baseline method that uses only color.}
\label{tab:davis_sota}
\end{table}

\paragraph{Limitations and further scope of improvement}
Compared to DAVIS (in Table \ref{tab:davis_sota}), our performance deteriorated for YouTube VOS (in Table \ref{tab:uvos_sota}). We observe that two primary reasons lead to a decline in performance. First, the object re-identification network \cite{li2017video} from \cite{premvos} has been used as part of our tracking pipeline is trained on DAVIS. Hence, most of the scenarios where objects undergo a change in appearance in the next consecutive frames have failed to re-identify the proposals from the previous frame. Second, video sequences from YouTube-VOS are relatively long compared to DAVIS and suffer more occlusions. 
\begin{figure}[!htp]
    \centering
    \includegraphics[width=1\linewidth]{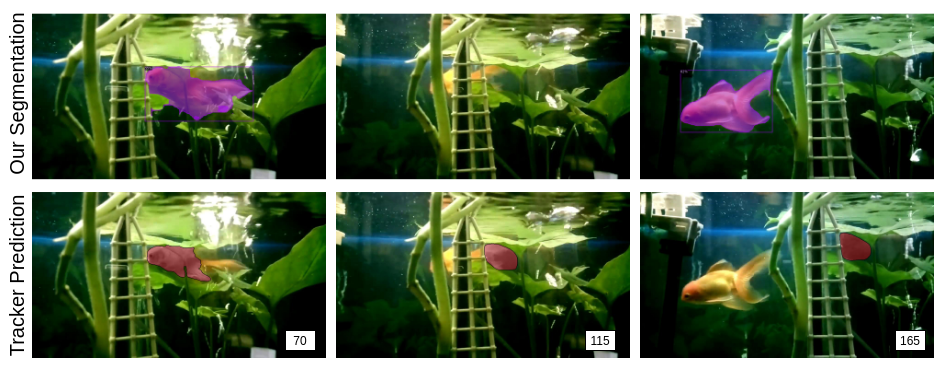}
    \caption{An example of negative scenario: tracking failure in YouTube VOS. The fish is present in our segmentation proposal but the tracker could not track it after it was occluded.}
    \label{fig:limitation}\vspace{-1em}
\end{figure}
Figure \ref{fig:limitation}, shows the initial appearance of fish, its occlusion, and its re-appearance from occlusion alongside our segmentation proposal and tracker prediction. We can observe that the tracker failed to re-identify despite having the proposal after the fish appeared from occlusion. This limitation can be addressed in two ways. First, the re-identification network can be fine-tuned with YouTube-VOS, which will help to re-identify objects in consecutive frames. Second, state-of-the-art tracking such as \cite{Xie} can be employed with a more extended temporal matching scheme that can effectively track the occluded object and effectively re-identify them even after the occlusion. Due to the scope of this work on box-supervised VOS, we will address these issues in our future work.

\begin{table}[htp]
\center
\scriptsize
\begin{tabular}{ccccc}
\toprule \multirow{2}{*}{ Method } & \multirow{2}{*}{ Sup. } & \multirow{2}{*}{ Overall $\uparrow$} & \multicolumn{2}{c}{ Seen } \\
\cline { 4 - 5 } & & & $\mathcal{J} \uparrow$ & $\mathcal{F} \uparrow$ \\\midrule \midrule
Vid. Color.\cite{vid_col} & Self & $38.9$ & $43.1$ & $38.6$ \\
CorrFlow\cite{CorrFlow} & Self & $46.6$ & $50.6$ & $46.6$  \\
\midrule

BoxInst\cite{Tian2020} $\dagger$ & Box & $48.2$ & $51.6$ & $52.3$  \\
\textbf{Ours} & \textbf{Box} & $\mathbf{53.5}$ & $\mathbf{58.7}$ & $\mathbf{59.2}$  \\ \midrule

OSMN\cite{osmn} & Full & $51.2$ & $60.0$ & $60.1$  \\
MSK\cite{msk} & Full & $53.1$ & $59.9$ & $59.5$   \\
RGMP\cite{rgmp} & Full & $53.8$ & $59.5$ & $-$   \\
OnAVOS\cite{onavos} & Full & $55.2$ & $60.1$ & $62.7$ \\
RVOS\cite{rvos} & Full & $56.8$ & $63.6$ & $67.2$  \\
OSVOS\cite{osvos} & Full & $58.8$ & $59.8$ & $60.5$  \\
S2S\cite{s2s} & Full & $64.4$ & $71.0$ & $70.0$  \\
\textcolor{violet}{PReMVOS\cite{premvos}} & \textcolor{violet}{Full} & \textcolor{violet}{$66.9$} & \textcolor{violet}{$71.4$} & \textcolor{violet}{$75.9$}   \\
STM\cite{stm} & Full & $79.4$ & $79.7$ & $84.2$ \\
\textbf{RMNet\cite{Xie}} & \textbf{Full} & $\mathbf{81.5}$ & $\mathbf{82.1}$ & $\mathbf{85.7}$ \\ \hline
\end{tabular}
\caption{ State-of-the-art comparison on Youtube-VOS\cite{yvos2018}. 
$\dagger$ denotes our baseline method that uses only color. The evaluation for unseen categories are reported in the appendix.}
\label{tab:uvos_sota}
\end{table}

\begin{table*}
\begin{tabular}[t]{lll}
\parbox[t]{.30\linewidth}{
\begin{tabular}[t]{c|c}\toprule
Filters & $\mathcal{J}\& \mathcal{F}($Mean$)$ \\\midrule\midrule
None  & 68.5 \\
BF*  &69.1 \\
BF* + TM& 69.2 \\
BF* + BDF + TM &  70.5 \\
\bottomrule
\end{tabular}
\caption{ Impact of the proposed filters. BF*: Box Filter, BDF: Bi-Directional Filter, TM: Temporal matching} 
\label{tab:ablation_motion} 
} &
\parbox[t]{.30\linewidth}{
\begin{tabular}[t]{c|c}\toprule
Supervision  & $\mathcal{J}\& \mathcal{F}($Mean$)$ \\\midrule\midrule
Motion Only  &  69.4 \\
Color $\cup$ Motion  & 68.8\\
Color $\cap$ Motion  &  70.5 \\
\bottomrule
\end{tabular}
\caption{Effect of combined motion and color utilization on pseudo mask generation.}
\label{tab:ablation_fusion} 
} &
\parbox[t]{.30\linewidth}{
\begin{tabular}[t]{c|c|c}\toprule
Backbone & T & $ \mathcal{J}\& \mathcal{F}($Mean$)$ \\\midrule\midrule
ResNet-50 & D & 70.5\\
ResNet-50 & DY &  72.1 \\
\textbf{ResNet-101} & \textbf{DY} &  \textbf{72.5}\\
\bottomrule
\end{tabular}
\caption{Effect of various backbone networks and data used for training. T: Train Data, D: DAVIS, Y: YVOS}
\label{tab:ablation_backbone}
}
\end{tabular}
\end{table*}

\subsection{Ablation Study}
We investigate five aspects of our method: improvement due to motion, motion pipeline and affinity loss, pre-training, experimenting with predicted foreground pixels, and finally, frame-level evaluation. Unless otherwise specified, the ablation study has been conducted using ResNet-50 \cite{he2016deep} as the backbone and trained with the DAVIS training dataset for faster computation.

\paragraph{Motion} Table \ref{tab:ablation_motion}, shows the influence of different motion filters and their combination as discussed in Sec. \ref{sec:motion_calc}. Unfiltered motion is inherently noisy and hurts the performance, whereas adding more filters for refinement of motion improves the performance. Starting from the box filter (BF) to the temporal matching (TM)  has incrementally refined motion and improved the segmentation performance. \\
Table \ref{tab:ablation_fusion}, describes the impact of the combination of color and motion during supervision as stated in Sec. \ref{sec:loss}. Here, the third row is the intersection affinity as in Eq. \ref{eq:aff_i} which is replaced with a union operation in the first row's experiment. Taking the union of color and motion supervision performed worse than when motion was used alone. However, the intersection of motion and color similarity gives the best performance in comparison to the union, hence it bolsters our strategy on generating small yet high precision foreground pixels. 
\paragraph{Backbone and Data} Finally, Table \ref{tab:ablation_backbone}, depicts the type of input backbone and data-sets used during training. A larger backbone network and more data help to learn more expressive features and improve performance.

\paragraph{Predicted Foreground pixels } Table \ref{tab:ablation-gt}, shows a proof-of-concept for the inclusion of a filter-based motion pipeline and its contribution to generating high-precision pixels. To calculate metrics (i.e., true positives) related to F1 score, precision, and recall, ground truth segmentation masks were used here. However, ground truth masks have never been used in training or performing filter optimization. Pixels greater than the threshold value of 0.5 are labeled as foreground, with the motion being normalized between 0 and 1. 
\begin{table}[!htp]\centering
\scriptsize
\begin{tabular}{c|c|c|c|c}
\toprule
{Filters} & {F1} & {Precision} & {Recall} & {Avg. FPS}\\
\midrule \midrule
None & 0.404 & 0.318 & 0.553 & \textbf{9.99}\\
BF* & \textbf{0.482} & 0.397 & \textbf{0.612} & 4.05 \\
BF* + TM + BDF & 0.460 & \textbf{0.421} & 0.507 & 1.96\\

\bottomrule
\end{tabular}
    \begin{tablenotes}
        \smaller
        \item 
        \begin{center}
            BF*: Box Filter, BDF: Bi-Directional Filter, TM: Temporal matching, FPS: Frame per second
        \end{center}
    \end{tablenotes}
\caption{ The capability of different filter combinations to extract foreground pixels for DAVIS trainval data-set.}
\label{tab:ablation-gt}
\end{table}
It can be observed from Table \ref{tab:ablation-gt} that using raw motion results in low precision, which is a deteriorating factor for our method. On the contrary, refined motion increases precision, resulting in improved quality of supervision and performance.

\paragraph{Frame level evaluation:}
To solely understand the proposed video segmentation proposal network, we measure frame-wise mAP. This estimates the true performance gain in frame-wise mask generation if we combine motion and color information. Table \ref{tab:frame_level}, show the performance of our method in comparison with weak box and full mask supervision. For Youtube-VOS, we hold out 20\% training data (YVOS-train-val) for evaluating frame-level segmentation as its ground truth validation masks are not publicly available. 
Table \ref{tab:frame_level}, clearly shows that our method consistently improves the quality of frame-wise mask compared to our baseline trained with only color information.  
\begin{table}[!htp]\centering
\scriptsize
\begin{tabular}{c|c|c|c}\toprule
Sup & Method & \multicolumn{2}{c}{ mAP } \\
\cline { 3-4 } & & YVOS-train-val &  DAVIS-val \\
\midrule
Box & BoxInst \cite{Tian2020} $\dagger$ & 31.3  & 24.2  \\
Box & Ours & 34  &  28.4  \\
Full & BoxInst with Mask Annotation $\dagger$ & 41.8 & 35.2 \\

\bottomrule
\end{tabular}
\caption{Performance evaluation on frame level segmentation proposal generation. All methods have the same network architecture (R-50 backbone) and similar training schedule.  $\dagger$ denotes that network is only supervised with only color information.}
\label{tab:frame_level}
\end{table}


\section{Conclusion}
Our work is the first to explore the potential of a motion-aware, box-supervised video segmentation proposal network. Furthermore, we have demonstrated that exploring video cues in a weakly supervised setup could develop a competitive VOS framework. The core idea is to refine and leverage motion and, subsequently, combine it with color channels if both share similar characteristics. Our proposed method significantly reduces the performance gap with top-performing fully supervised methods on the DAVIS dataset. At the same time, we are addressing some of the inherent complex scenarios in VOS, such as the camouflage effect. We will investigate a more accurate motion compensation pipeline and an extended temporal proposal tracking for an efficient VOS framework as part of future work. We sincerely hope our weak box supervised video object segmentation work will pave the way for new VOS research.

\color{black}
\newpage
{\small
\bibliographystyle{ieee}
\bibliography{egbib}
}
\clearpage

\newpage
\appendix
\onecolumn
\section{Appendix.}

\subsection{Dataset Preparation}
To train the segmentation network on YoutubeVOS \cite{yvos2018} and DAVIS \cite{davis17} datasets, we first created COCO \cite{coco} style annotations and trained the model with independent frames. We computed the bounding box from the ground truth segmentation mask as no bounding box is available in the datasets. The ground truth mask is not used for any other training purposes. 

\subsection{Training Procedure}
We have implemented our method on the detectron2 \cite{wu2019detectron2} or more specifically its adapted version, AdelaiDet \cite{tian2019adelaidet} framework. Our segmentation network is first trained on the YoutubeVOS dataset and then fine-tuned on the DAVIS dataset. We trained the model with the same configuration similar to CondInst \cite{tian2020conditional} with some changes in hyperparameters as listed below. A more detailed configuration can be found in our code. We define one batch as an iteration.

\paragraph{YoutubeVOS training:}
We trained our segmentation network on the YoutubeVOS dataset using an SGD optimizer with a gradual warmup till 10000 iterations and batch size of 12. The base learning rate is set to 0.01, and further, it reduces by 0.1 factor after  60000 and 80000 iterations. We train till 100K iterations. We used the ResNet-101 as a backbone feature extractor and ran it on three Quadro RTX 8000 GPU machines.

\paragraph{DAVIS Fine-tuning.}
For finetuning on the DAVIS dataset, we set the learning rate very low at 0.001. We evaluate our model after 350 iterations and report our best model at 6300 iterations.

\subsection{More on Qualitative Study}
\begin{figure}[!htp]
\centering
\includegraphics[width=\linewidth]{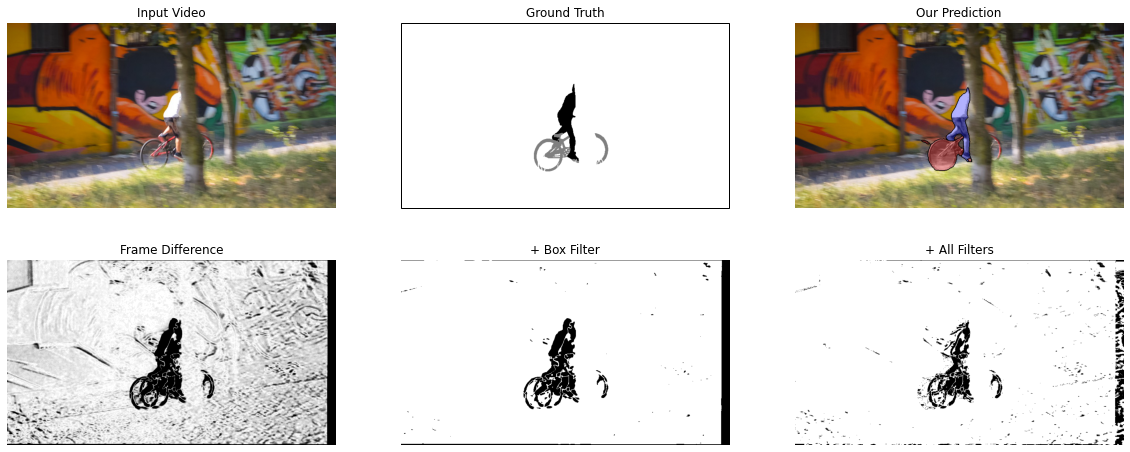}
\caption{\textbf{An example frame from the video sequence.} First column represents side-by-side portraying of input, ground truth and prediction. Moreover the second column shows proposed sequential improvement of motion map.}
\label{fig:example}
\end{figure}

In addition to the qualitative example presented in the main paper, we have created a video containing five positive and one negative example. Similar to Figure \ref{fig:example}, a sample video contained the input frames, ground truth, various motion maps, and predicted masks for better comparisons. One such example can be seen in Fig. \ref{fig:example}

\newpage
\subsection{The improvement from motion supervision}.
\begin{figure}[h]
    \centering
    \includegraphics[width=0.75\linewidth]{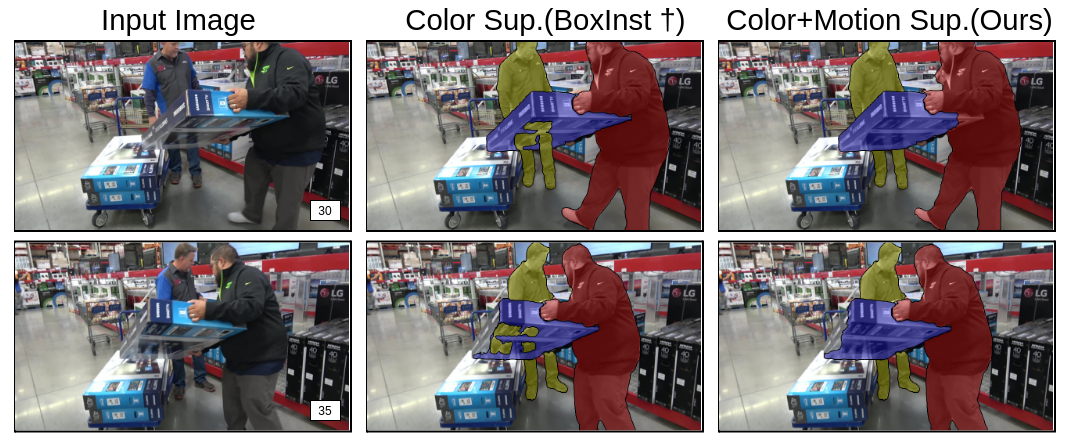}
    \caption{A qualitative comparison between baseline(color only) and our method. The baseline model predicts an erroneous mask for the distant man due to color similarity with reflection on the TV case. While the inclusion of motion alleviates this in our method.}
    \label{fig:c_v_m}
\end{figure}

\subsection{Improvement with Temporal Matching}
\begin{figure}[!h]
    \centering
    \includegraphics[width=0.75\linewidth]{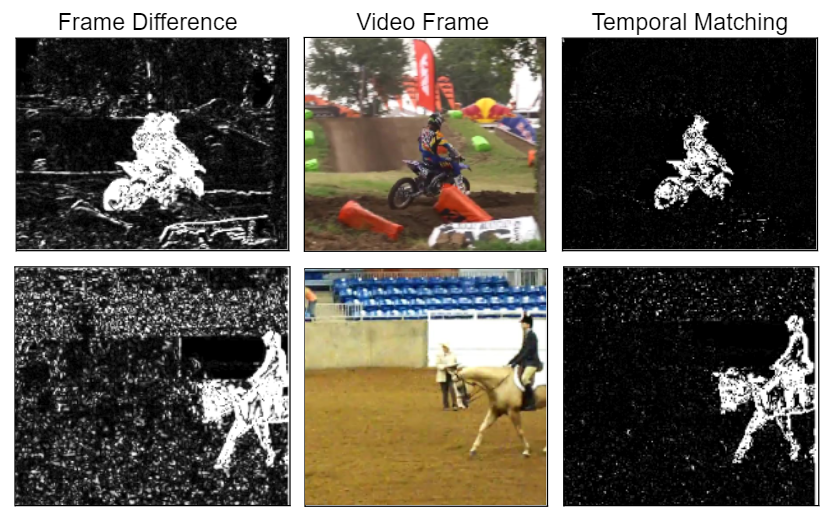}
    \caption{Examples of improved motion map with Temporal Matching.}
    \label{fig:three_frame}
\end{figure}

The fast-moving bike in Figure \ref{fig:three_frame}A, seems to appear twice. Whereas, in Figure \ref{fig:three_frame}D, the horse leg has thick edges due to its slow movement. The Temporal matching solved these issues.

\newpage

\subsection{Performance comparison for Unseen classes of Youtube-VOS dataset.}
\begin{table}[!htp]
\centering
\begin{tabular}{ccccc}
\toprule \multirow{2}{*}{ Method } & \multirow{2}{*}{ Sup. } & \multirow{2}{*}{ Overall $\uparrow$}  & \multicolumn{2}{c}{ Unseen } \\
\cline { 4 - 5 } & & & $\mathcal{J} \uparrow$ & $\mathcal{F} \uparrow$ \\\midrule \midrule
Vid. Color.\cite{vid_col} & Self & $38.9$ & $36.6$ & $37.4$ \\
CorrFlow\cite{CorrFlow} & Self & $46.6$ & $43.8$ & $45.6$ \\
\midrule

BoxInst\cite{Tian2020} $\dagger$ & Box & $48.2$ & $44.7.8$ & $44.9$ \\
\textbf{Ours} & \textbf{Box} & $\mathbf{53.5}$ &  $\mathbf{46.3}$ & $\mathbf{48.4}$ \\ \midrule

OSMN\cite{osmn} & Full & $51.2$ & $40.6$ & $44.0$ \\
MSK\cite{msk} & Full & $53.1$ & $45.0$ & $47.9$ \\
RGMP\cite{rgmp} & Full & $53.8$ & $45.2$ & $-$ \\
OnAVOS\cite{onavos} & Full & $55.2$ & $46.6$ & $51.4$ \\
RVOS\cite{rvos} & Full & $56.8$ & $45.5$ & $51.0$ \\
OSVOS\cite{osvos} & Full & $58.8$ & $54.2$ & $60.7$ \\
S2S\cite{s2s} & Full & $64.4$ & $55.5$ & $61.2$ \\
\textcolor{violet}{PReMVOS\cite{premvos}} & \textcolor{violet}{Full} & 
\textcolor{violet}{$66.9$} &
\textcolor{violet}{$56.5$} & \textcolor{violet}{$63.7$} \\
STM\cite{stm} & Full & $79.4$ & $72.8$ & $80.9$ \\
\textbf{RMNet\cite{Xie}} & \textbf{Full} & $\mathbf{81.5}$ & $\mathbf{75.7}$ & $\mathbf{82.4}$ \\ \hline
\end{tabular}
\caption{ State-of-the-art comparison on Youtube-VOS\cite{yvos2018} on unseen categories. 
$\dagger$ denotes our baseline method that uses only color.}
\label{tab:uvos_sota}
\end{table}

\end{document}